\documentclass[11pt,a4paper]{article}
\usepackage[hyperref]{naaclhlt2019}
\usepackage{latexsym,booktabs,multicol,multirow}
\usepackage{times}
\usepackage{graphicx}
\usepackage[T1,T5]{fontenc}
\usepackage[utf8]{inputenc}
\usepackage{tikz}
\usepackage{amsmath}
\usepackage{xcolor}
\usepackage{collcell}
\usepackage{float}
\usepackage{natbib}
\usepackage[font={small}]{caption}
\usepackage{colortbl,dcolumn}
\usepackage{array}

\usepackage{url}

\aclfinalcopy 

\usepackage{arydshln} 

\title{Recursive Subtree Composition in LSTM-Based Dependency Parsing}

\author{Miryam de Lhoneux$^{\spadesuit}$~~~Miguel Ballesteros$^{\diamondsuit}$~~~Joakim Nivre$^{\spadesuit}$ \\
$^{\spadesuit}$ Department of Linguistics and Philology, Uppsala University \\
$^{\diamondsuit}$ IBM Research AI, Yorktown Heights, NY \\
{\small \tt \{miryam.de\_lhoneux,joakim.nivre\}@lingfil.uu.se}\\ {\small \tt miguel.ballesteros@ibm.com}
}

\date{}

\begin{document}
\maketitle
\begin{abstract}
    The need for tree structure modelling on top of sequence modelling is an open issue in neural dependency parsing.
    We investigate the impact of adding a tree layer on top of a sequential model by recursively composing subtree representations (composition) in a transition-based parser that uses features extracted by a \mbox{BiLSTM}. Composition seems superfluous with such a model, suggesting that \mbox{BiLSTMs} capture information about subtrees. We perform model ablations to tease out the conditions under which composition helps. When ablating the backward LSTM, performance drops and composition does not recover much of the gap. When ablating the forward LSTM, performance drops less dramatically and composition recovers a substantial part of the gap, indicating that a forward LSTM and composition capture similar information. We take the backward LSTM to be related to lookahead features and the forward LSTM to the rich history-based features both crucial for transition-based parsers. To capture history-based information, composition is better than a forward LSTM on its own, but it is even better to have a forward LSTM as part of a \mbox{BiLSTM}. We correlate results with language properties, showing that the improved lookahead of a backward LSTM is especially important for head-final languages.
\end{abstract}

\section{Introduction}
Recursive neural networks allow us to construct vector representations of trees or subtrees. They have been used for constituency parsing by \citet{socher13parsing} and \citet{dyer16recurrent} and for dependency parsing by \citet{stenetorp2013transition} and \citet{dyer15}, among others.
In particular, \citet{dyer15} showed that composing representations of subtrees using recursive neural networks can be beneficial for transition-based dependency parsing. These results were further strengthened in \citet{kuncoro17recurrent} who showed, using ablation experiments, that composition is key in the Recurrent Neural Network Grammar (RNNG) generative parser by \citet{dyer16recurrent}.

In a parallel development, \citet{kiperwasser16} showed that using \mbox{BiLSTMs} for feature extraction can lead to high parsing accuracy even with fairly simple parsing architectures, and using \mbox{BiLSTMs} for feature extraction has therefore become very popular in dependency parsing. It is used in the state-of-the-art parser of \citet{dozat16biaffine}, was used in 8 of the 10 highest performing systems of the 2017 CoNLL shared task \citep{udst:overview} and 10 out of the 10 highest performing systems of the 2018 CoNLL shared task \citep{conll2018}.

This raises the question of whether features extracted with \mbox{BiLSTMs} in themselves capture information about subtrees, thus making recursive composition superfluous. Some support for this hypothesis comes from the results of \citet{linzen16assessing} which indicate that LSTMs can capture hierarchical information: they can be trained to predict long-distance number agreement in English. Those results were extended to more constructions and three additional languages by \citet{gulordava18colorless}. 
However, \citet{kuncoro2018lstms} have also shown that although sequential LSTMs can learn syntactic information, a recursive neural network which explicitly models hierarchy (the RNNG model from \citet{dyer15}) is better at this: it performs better on the number agreement task from \citet{linzen16assessing}. 

To further explore this question in the context of dependency parsing, we investigate the use of recursive composition (henceforth referred to as \emph{composition}) in a parser with an architecture like the one in \citet{kiperwasser16}. This allows us to explore variations of features and isolate the conditions under which composition is helpful. We hypothesise that the use of a \mbox{BiLSTM} for feature extraction makes it possible to capture information about subtrees and therefore makes the use of subtree composition superfluous. We hypothesise that composition becomes useful when part of the BiLSTM is ablated, the forward or the backward LSTM. We further hypothesise that composition is most useful when the parser has no access to information about the function of words in the context of the sentence given by POS tags. When using POS tags, the tagger has indeed had access to the full sentence. We additionally look at what happens when we ablate character vectors which have been shown to capture information which is partially overlapping with information from POS tags.
We experiment with a wider variety of languages than \citet{dyer15} in order to explore whether the usefulness of different model variants vary depending on language type.

\section{K\&G Transition-Based Parsing}
\indent We define the parsing architecture introduced by \citet{kiperwasser16} at a high level of abstraction and henceforth refer to it as K\&G.
A K\&G parser is a greedy transition-based parser.\footnote{\citet{kiperwasser16} also define a graph-based parser with similar feature extraction, but we focus on transition-based parsing.} 
For an input sentence of length $n$ with words $w_1,\dots,w_n$, a sequence of vectors $x_{1:n}$ is created, where the vector $x_i$ is a vector representation of the word $w_i$. We refer to these as \emph{type} vectors, as they are the same for all occurrences of a word type. Type vectors are then passed through a feature function which learns representations of words in the context of the sentence.
\begin{align*}
    x_i &= e(w_i)\\
    v_i &= f(x_{1:n},i)
\end{align*}
We refer to the vector $v_i$ as a \emph{token} vector, as it is different for different tokens of the same word type. In \citet{kiperwasser16}, the feature function used is a \mbox{BiLSTM}.\\
\indent As is usual in transition-based parsing, parsing involves taking transitions from an initial configuration to a terminal one. Parser configurations are represented by a stack, a buffer and set of dependency arcs \citep{nivre08cl}.
For each configuration $c$, the feature extractor concatenates the token representations of core elements from the stack and buffer. These token vectors are passed to a classifier, typically a Multilayer Perceptron (MLP). The MLP scores transitions together with the 
arc labels for transitions that involve adding an arc. Both the word type vectors and the \mbox{\mbox{BiLSTMs}} are trained together with the model.

\section{Composing Subtree Representations}
\indent \citet{dyer15} looked at the impact of using a recursive composition function in their parser, which is also a transition-based parser but with an architecture different from K\&G. 
They make use of a variant of the LSTM called a stack LSTM. A stack LSTM has push and pop operations which allow passing through states in a tree structure rather than sequentially. Stack LSTMs are used to represent the stack, the buffer, and the sequence of past parsing actions performed for a configuration.\\
\indent The words of the sentence are represented by vectors of the word types, together with a vector representing the word's POS tag. In the initial configuration, the vectors of all words are in the buffer and the stack is empty. The representation of the buffer is the end state of a backward LSTM over the word vectors. 
As parsing evolves, the word vectors are popped from the buffer, pushed to and popped from the stack and the representations of stack and buffer get updated.\\
\indent \citet{dyer15} define a recursive composition function and compose tree representations incrementally, as dependents get attached to their head. The composed representation $c$ is built by concatenating the vector $h$ of the head with the vector of the dependent $d$
, as well as a vector $r$ representing the label paired with the direction of the arc. That concatenated vector is passed through an affine transformation and then through a $tanh$ non-linear activation.
\begin{align*}
    c = tanh(W[h;d;r]+b)
\end{align*}

\noindent
They create two versions of the parser. In the first version, when a dependent is attached to a head, the word vector of the head is replaced by a composed vector of the head and dependent. In the second version, they simply keep the vector of the head when attaching a dependent to a head. They observe that the version with composition is substantially better than the version without, by 1.3 LAS points for English (on the Penn Treebank (PTB) test set) and 2.1 for Chinese (on the Chinese Treebank (CTB) test set).\\ 
\indent Their parser uses POS tag information. POS tags help to disambiguate between different functional uses of a word and in this way give information about the use of the word in context. We hypothesise that the effect of using a recursive composition function is stronger when not making use of POS tags.

\section{Composition in a K\&G Parser}
\indent The parsing architectures of the stack LSTM parser (S-LSTM) and K\&G are different but have some similarities.\footnote{Note that we use S-LSTM to denote the stack LSTM parser, not the stack LSTM as an LSTM type.} 
In both cases, the configuration is represented by vectors obtained by LSTMs. In K\&G, it is represented by the token vectors of top items of the stack and the first item of the buffer. In the S-LSTM, it is represented by the vector representations of the entire stack, buffer and sequence of past transitions.\\
\indent Both types of parsers learn vector representations of word types which are passed to an LSTM. 
In K\&G, they are passed to an LSTM in a feature extraction step that happens before parsing. The LSTM in this case is used to learn vectors that have information about the context of each word, a token vector. In the S-LSTM, word type vectors are passed to Stack LSTMs as parsing evolves. In this case, LSTMs are used to learn vector representations of the stack and buffer (as well as one which learns a representation of the parsing action history).\\
\indent When composition is not used in the S-LSTM, word vectors represent word types. When composition is used, as parsing evolves, the stack and buffer vectors get updated with information about the subtrees they contain, so that they gradually become contextualised. In this sense, those vectors become more like token vectors in K\&G. More specifically, as explained in the previous section, when a dependent is attached to its head, the composition function is applied to the vectors of head and dependent and the vector of the head is replaced by this composed vector. \\
\indent We cannot apply composition on type vectors in the K\&G architecture, since they are not used after the feature extraction step and hence cannot influence the representation of the configuration. Instead, we apply composition on the token vectors. We embed those composed representations in the same space as the token vectors.\\
\indent In K\&G, like in the S-LSTM, we can create a composition function and compose the representation of subtrees as parsing evolves. We create two versions of the parser, one where word tokens are represented by their token vector. The other where they are represented by their token vector and the vector of their subtree $c_i$, which is initially just a copy of the token vector ($v_i = f(x_{1:n},i)\circ c_i$). When a dependent word $d$ is attached to a word $h$ with a relation and direction $r$, $c_i$ is computed with the same composition function as in the S-LSTM defined in the previous section, repeated below.\footnote{Note that, in preliminary experiments, we tried replacing the vector of the head by the vector of its subtree instead of concatenating the two but concatenating gave much better results.} \\
\indent This composition function is a simple recurrent cell. Simple RNNs have known shortcomings which have been addressed by using LSTMs, as proposed by \citet{hochreiter1997long}. A natural extension to this composition function is therefore to replace it with an LSTM cell. We also try this variant. We construct LSTMs for subtrees. We initialise a new LSTM for each new subtree that is formed, that is, when a dependent $d$ is attached to a head $h$ which does not have any dependent yet. Each time we attach a dependent to a head, we construct a vector which is a concatenation of $h$, $d$ and $r$. We pass this vector to the LSTM of $h$. $c$ is the output state of the LSTM after passing through that vector.
We denote those models with $+rc$ for the one using an ungated recurrent cell and with $+lc$ for the one using an LSTM cell.
\vspace{-0.15cm}
\begin{align*}
    c &= tanh(W[h;d;r]+b)\\
    c &= \textsc{Lstm}([h;d;r])
\end{align*}

\noindent 
As results show (see §~\ref{sec:res}), neither type of composition seems useful when used with the K\&G parsing model, which indicates that \mbox{BiLSTMs} capture information about subtrees. To further investigate this and in order to isolate the conditions under which composition is helpful, we perform different model ablations and test the impact of recursive composition on these ablated models.\\
\indent First, we ablate parts of the \mbox{BiLSTMs}: we ablate either the forward or the backward LSTM. We therefore build parsers with 3 different feature functions $f(x, i)$ over the word type vectors $x_i$ in the sentence $x$: a \mbox{BiLSTM} ($bi$) (our baseline), a backward LSTM ($bw$) (i.e., ablating the forward LSTM) and a forward LSTM ($fw$) (i.e., ablating the backward LSTM):
\begin{align*}
    bi(x,i) &= \textsc{BiLstm}(x_{1:n},i) \\
    bw(x,i) &= \textsc{Lstm}(x_{n:1},i) \\
    fw(x,i) &= \textsc{Lstm}(x_{1:n},i) 
\end{align*}

\noindent K\&G parsers with unidirectional LSTMs are, in some sense, more similar to the S-LSTM than those with a \mbox{BiLSTM}, since the S-LSTM only uses unidirectional LSTMs. We hypothesise that composition will help the parser using unidirectional LSTMs in the same way it helps an S-LSTM. 

\indent We additionally experiment with the vector representing the word at the input of the LSTM. The most complex representation consists of a concatenation of an embedding of the word type $e(w_i)$, an embedding of the (predicted) POS tag of $w_i$, $p(w_i)$ and a character representation of the word obtained by running a BiLSTM over the characters $ch_{1:m}$ of $w_i$ ($\text{BiLSTM}(ch_{1:m})$).

\vspace{-0.15cm}
\begin{align*}
    x_i = e(w_i) \circ p(w_i) \circ \text{BiLSTM}(ch_{1:m})
\end{align*}

Without a POS tag embedding, the word vector is a representation of the word type. With POS information, we have some information about the word in the context of the sentence and the tagger has had access to the full sentence. The representation of the word at the input of the \mbox{BiLSTM} is therefore more contextualised and it can be expected that a recursive composition function will be less helpful than when POS information is not used.
Character information has been shown to be useful for dependency parsing first by \citet{ballesteros15}. \citet{ballesteros15} and \citet{smith18investigation} among others have shown that POS and character information are somewhat complementary. \citet{ballesteros15} used similar character vectors in the S-LSTM parser but did not look at the impact of composition when using these vectors. Here, we experiment with ablating either or both of the character and POS vectors. We look at the impact of using composition on the full model as well as these ablated models. We hypothesise that composition is most helpful when those vectors are not used, since they give information about the functional use of the word in context.

\paragraph*{Parser}
We use UUParser, a variant of the K\&G transition-based parser that employs the arc-hybrid transition system from \citet{kuhlmann11} extended with a \textsc{Swap} transition and a Static-Dynamic oracle, as described in \citet{delhoneux17arc}\footnote{The code can be found at \url{https://github.com/mdelhoneux/uuparser-composition}}. The \textsc{Swap} transition is used to allow the construction of non-projective dependency trees \citep{nivre09acl}. 
We use default hyperparameters. When using POS tags, we use the universal POS tags from the UD treebanks which are coarse-grained and consistent across languages. Those POS tags are predicted by UDPipe \citep{udpipe} both for training and parsing. This parser obtained the 7th best LAS score on average in the 2018 CoNLL shared task \citep{conll2018}, about 2.5 LAS points below the best system, which uses an ensemble system as well as ELMo embeddings, as introduced by \citet{elmo}. 
Note, however, that we use a slightly impoverished version of the model used for the shared task which is described in \citet{smith18treebanks}: we use a less accurate POS tagger (UDPipe) and we do not make use of multi-treebank models. In addition, \citet{smith18treebanks} use the three top items of the stack as well as the first item of the buffer to represent the configuration, while we only use the two top items of the stack and the first item of the buffer. \citet{smith18treebanks} also use an \textit{extended feature set} as introduced by \citet{kiperwasser16} where they also use the rightmost and leftmost children of the items of the stack and buffer that they consider. We do not use that extended feature set. This is to keep the parser settings as simple as possible and avoid adding confounding factors. It is still a near-SOTA model. We evaluate parsing models on the development sets and report the average of the 5 best results in 30 epochs and 5 runs with different random seeds. 

\paragraph*{Data}
\indent We test our models on a sample of treebanks from Universal Dependencies v2.1 \citep{udv21}. We follow the criteria from \citet{delhoneux17old} to select our sample: we ensure typological variety, we ensure variety of domains, we verify the quality of the treebanks, and we use one treebank with a large amount of non-projective arcs. However, unlike them, we do not use extremely small treebanks. Our selection is the same as theirs but we remove the tiny treebanks and replace them with 3 others. Our final set is: Ancient Greek (PROIEL), Basque, Chinese, Czech, English, Finnish, French, Hebrew and Japanese. 

\section{Results}
\label{sec:res}
\begin{figure}[!ht]
    \centering
    \includegraphics[scale=0.44]{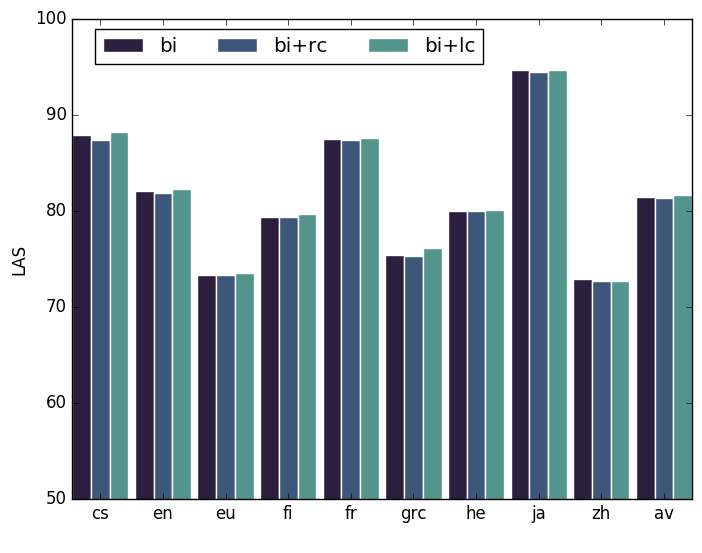}
    \caption{\label{fig:compos_bi}LAS of models using a BiLSTM ($bi$) without composition, with a recurrent cell ($+rc$) and with an LSTM cell ($+lc$). Bar charts truncated at 50 for visualization purposes.}
\end{figure}

\indent First, we look at the effect of our different recursive composition functions on the full model (i.e., the model using a BiLSTM for feature extraction as well as both character and POS tag information). As can be seen from Figure~\ref{fig:compos_bi}, recursive composition using an LSTM cell ($+lc$) is generally better than recursive composition with a recurrent cell ($+rc$), but neither technique reliably improves the accuracy of a \mbox{BiLSTM} parser.

\subsection{Ablating the forward and backward LSTMs}

\begin{figure}
   \centering
    \includegraphics[scale=0.4]{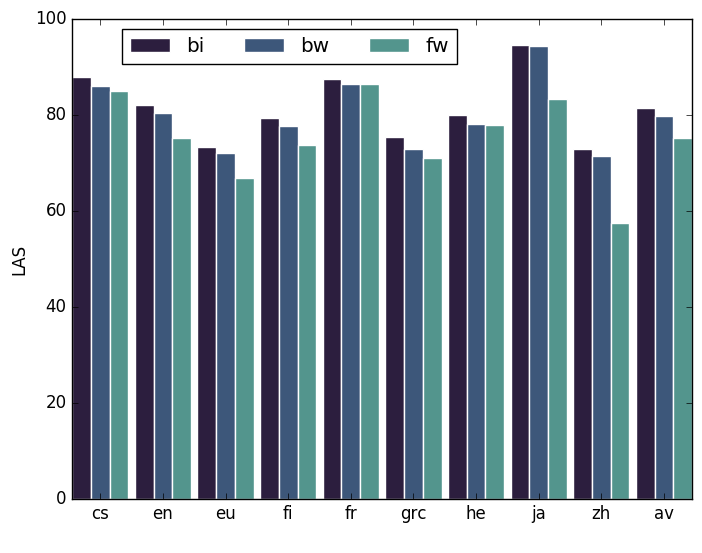}
    \caption{\label{fig:bibwfw}LAS of models using a BiLSTM ($bi$), backward LSTM ($bw$) and forward LSTM ($fw$).}
\end{figure}

\begin{figure}
    \includegraphics[scale=0.30]{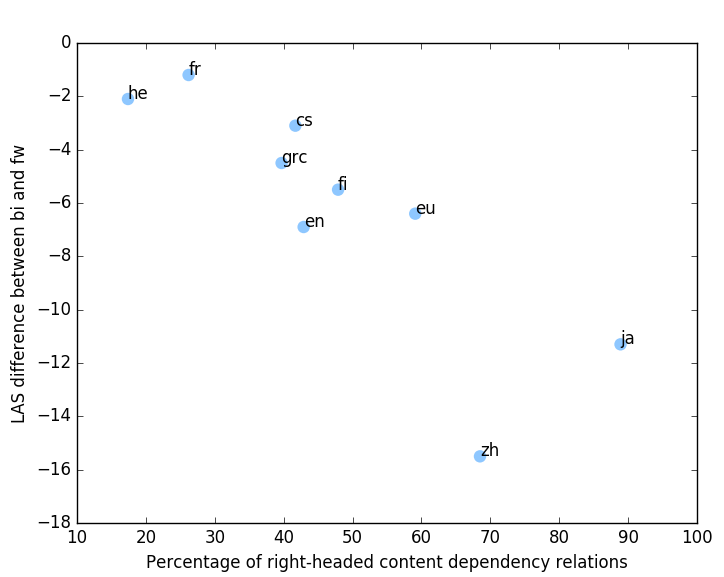}
    \caption{\label{fig:correl} Correlation between how hurtful it is to ablate the backward LSTM and right-headedness of languages. }
\end{figure}

\indent Second, we only consider the models using character and POS information and look at the effect of ablating parts of the \mbox{BiLSTM} on the different languages. The results can be seen in Figure~\ref{fig:bibwfw}. As expected, the \mbox{BiLSTM} parser performs considerably better than both unidirectional LSTM parsers, and the backward LSTM is considerably better than the forward LSTM, on average. It is, however, interesting to note that using a forward LSTM is much more hurtful for some languages than others: it is especially hurtful for Chinese and Japanese. This can be explained by language properties: the right-headed languages suffer more from ablating the backward LSTM 
than other languages. We observe a correlation between how hurtful a forward model is compared to the baseline and the percentage of right-headed content dependency relations ($R=-0.838$, $p<.01$), see Figure~\ref{fig:correl}.\footnote{The reason we only consider content dependency relations is that the UD scheme focuses on dependency relations between content words and treats function words as features of content words to maximise parallelism across languages \citep{demarneffe14}.} 

There is no significant correlation between how hurtful ablating the forward LSTM is and the percentage of left-headed content dependency relations ($p>.05$) indicating that its usefulness is not dependent on language properties. We hypothesise that dependency length or sentence length can play a role but we also find no correlation between how hurtful it is to ablate the forward LSTM and average dependency or sentence length in treebanks.
It is finally also interesting to note that the backward LSTM performance is close to the \mbox{BiLSTMs} performance for some languages (Japanese and French).

\begin{figure*}[!ht]
    \centering
    \includegraphics[scale=0.44]{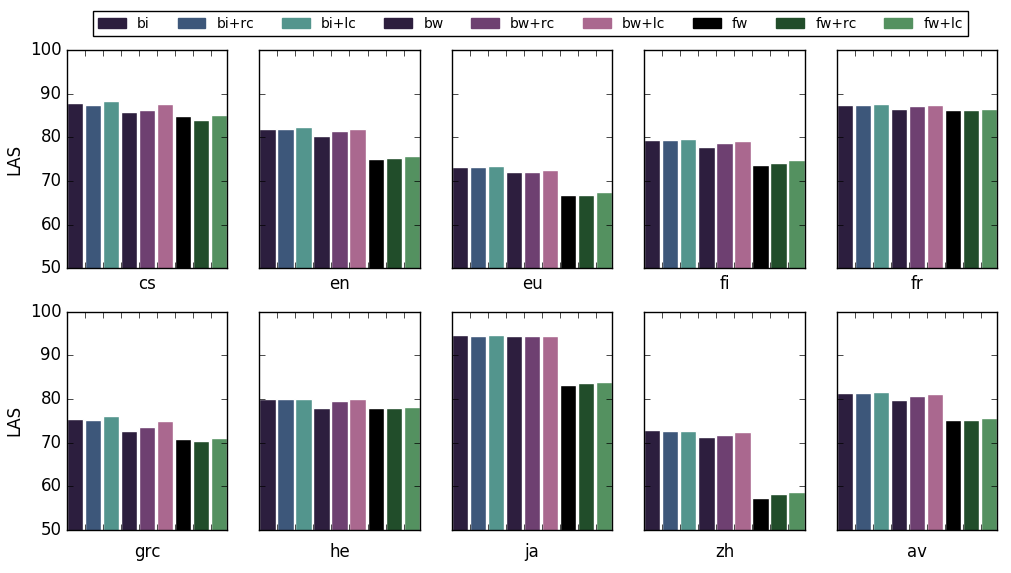}
    \caption{\label{fig:compos}LAS of models using a BiLSTM ($bi$), backward LSTM ($bw$) and forward LSTM ($fw$), without recursive composition, with a recurrent cell ($+rc$) and with a LSTM cell ($+lc$). Bar charts truncated at 50 for visualization purposes.}
\end{figure*}

We now look at the effect of using recursive composition on these ablated models. Results are given in Figure~\ref{fig:compos}. 
First of all, we observe unsurprisingly that composition using an LSTM cell is much better than using a simple recurrent cell. Second, both types of composition help the backward LSTM case, but neither reliably helps the $bi$ models. Finally, the recurrent cell does not help the forward LSTM case but the LSTM cell does to some extent. It is interesting to note that using composition, especially using an LSTM cell, bridges a substantial part of the gap between the $bw$ and the $bi$ models.

\indent These results can be related to the literature on transition-based dependency parsing. Transition-based parsers generally rely on two types of features: \emph{history-based features} over the emerging dependency tree and \emph{lookahead features} over the buffer of remaining input. The former are based on a hierarchical structure, the latter are purely sequential. \citet{mcdonald07emnlp} and \citet{mcdonald11cl} have shown that history-based features enhance transition-based parsers as long as they do not suffer from error propagation. However, \citet{nivre06book} has also shown that lookahead features are absolutely crucial given the greedy left-to-right parsing strategy.

In the model architectures considered here, the backward LSTM provides an improved lookahead. Similarly to the lookahead in statistical parsing, it is sequential. The difference is that it gives information about upcoming words with unbounded length. The forward LSTM in this model architecture provides history-based information but unlike in statistical parsing, that information is built sequentially rather than hierarchically: the forward LSTM passes through the sentence in the linear order of the sentence. In our results, we see that lookahead features are more important than the history-based ones. It hurts parsing accuracy more to ablate the backward LSTM than to ablate the forward one. This is expected given that some history-based information is still available through the top tokens on the stack, while the lookahead information is almost lost completely without the backward LSTM.\\
\indent A composition function gives hierarchical information about the history of parsing actions. It makes sense that it helps the backward LSTM model most since that model has no access to any information about parsing history. It helps the forward LSTM slightly which indicates that there can be gains from using structured information about parsing history rather than sequential information. We could then expect that composition should help the \mbox{BiLSTM} model which, however, is not the case. This might be because the \mbox{BiLSTM} constructs information about parsing history and lookahead into a unique representation. In any case, this indicates that \mbox{BiLSTMs} are powerful feature extractors which seem to capture useful information about subtrees.

\subsection{Ablating POS and character information}
\begin{figure}
	\centering
    \includegraphics[scale=0.4]{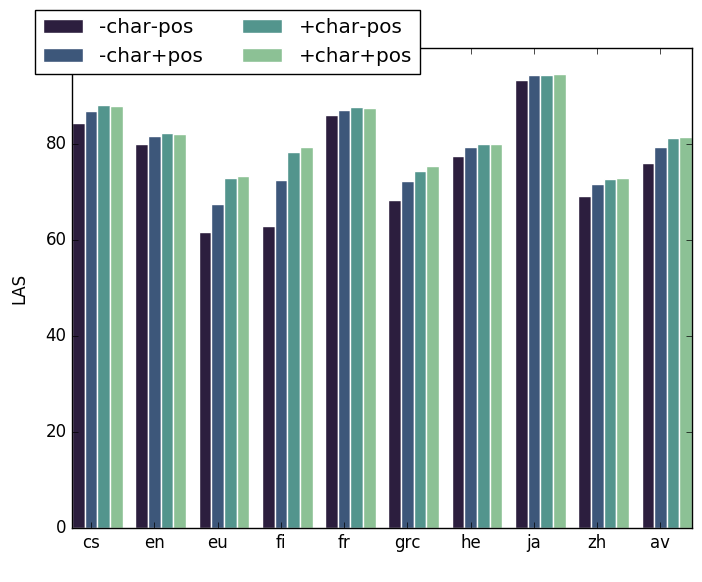}
    \caption{\label{fig:bi_feat}LAS of baseline, using char and/or POS tags to construct word representations}
\end{figure}

\indent Next, we look at the effect of the different word representation methods on the different languages, as represented in Figure~\ref{fig:bi_feat}. As is consistent with the literature \citep{ballesteros15,delhoneux17raw,smith18investigation}, using character-based word representations and/or POS tags consistently improves parsing accuracy but has a different impact in different languages and the benefits of both methods are not cumulative: using the two combined is not much better than using either on its own. In particular, character models are an efficient way to obtain large improvements in morphologically rich languages.

    \begin{table*}[t]
        \centering
        \small
        \begin{tabular}{l|ll:ll:ll|ll:ll:ll}
            \multicolumn{7}{c}{\bf{pos+char+}}                 & \multicolumn{6}{c}{\bf{pos+char-}}           \\
            \toprule
            & bi   & bi+lc & bw   & bw+lc & fw   & fw+lc & bi   & bi+lc & bw   & bw+lc & fw   & fw+lc \\
            \midrule
            cs  & 87.9 & 88.2          & 85.9 & \textbf{87.7} & 84.9 & 85.0          & 86.7      & 87.0          & 84.5 & \textbf{86.2} & 83.6 & 83.6          \\
            en  & 82.0 & 82.3          & 80.3 & \textbf{81.9} & 75.1 & \textbf{75.6} & 81.5      & 81.5          & 79.7 & \textbf{81.4} & 74.3 & \textbf{75.0}          \\
            eu  & 73.3 & 73.5          & 72.0 & 72.4          & 66.8 & \textbf{67.4} & 67.4      & 67.6          & 65.6 & \textbf{66.3} & 59.6 & \textbf{60.5}          \\
            fi  & 79.3 & 79.7          & 77.7 & \textbf{79.2} & 73.7 & \textbf{74.7} & 72.5      & 72.7          & 69.8 & \textbf{71.7} & 66.7 & \textbf{67.4}          \\
            fr  & 87.5 & 87.6          & 86.4 & \textbf{87.5} & 86.3 & 86.4          & 87.1      & 87.2          & 85.8 & \textbf{86.9} & 85.7 & 85.9          \\
            grc & 75.4 & \textbf{76.1} & 72.8 & \textbf{75.0} & 70.9 & 71.1          & 72.2      & 72.5          & 69.6 & \textbf{71.4} & 67.4 & 67.8          \\
            he  & 80.0 & 80.1          & 78.0 & \textbf{80.0} & 77.9 & 78.2          & 79.4      & 79.2          & 77.2 & \textbf{79.0} & 76.9 & 77.3          \\
            ja  & 94.6 & 94.6          & 94.4 & 94.5          & 83.3 & \textbf{83.9} & 94.3      & 94.3          & 94.2 & 94.3          & 83.0 & \textbf{83.6} \\
            zh  & 72.9 & 72.7          & 71.3 & \textbf{72.4} & 57.4 & \textbf{58.7} & 71.5      & 71.3          & 69.9 & \textbf{70.8} & 56.4 & \textbf{57.9} \\
            \midrule
            av  & 81.4 & 81.6          & 79.8 & \textbf{81.2} & 75.1 & \textbf{75.7} & 79.2      & 79.2          & 77.4 & \textbf{78.7} & 72.6 & \textbf{73.2} \\
            \bottomrule
        \end{tabular}
        \begin{tabular}{l|ll:ll:ll|ll:ll:ll}
            \multicolumn{7}{c}{\bf{pos-char+}}                 & \multicolumn{6}{c}{\bf{pos-char-}}           \\
            & bi   & bi+lc & bw   & bw+lc & fw   & fw+lc & bi   & bi+lc & bw   & bw+lc & fw   & fw+lc \\
            \midrule
            cs  & 88.1 & 88.4          & 86.0 & \textbf{87.8} & 84.7 & 84.9          & 84.3      & 84.5          & 81.3 & \textbf{83.1} & 79.9 & 79.8          \\
            en  & 82.2 & 82.1          & 79.8 & \textbf{81.6} & 73.2 & \textbf{73.8} & 80.0      & 79.9          & 77.5 & \textbf{79.2} & 70.5 & \textbf{71.5} \\
            eu  & 72.8 & 72.9          & 71.5 & 71.8          & 65.4 & \textbf{66.4} & 61.6      & 62.0          & 57.7 & \textbf{59.5} & 48.7 & \textbf{51.2} \\
            fi  & 78.2 & 78.6          & 75.8 & \textbf{77.9} & 72.0 & \textbf{73.0} & 62.8      & 63.1          & 56.6 & \textbf{60.2} & 52.8 & \textbf{54.7} \\
            fr  & 87.6 & 87.7          & 86.1 & \textbf{87.4} & 85.4 & 85.7          & 85.9      & 85.8          & 83.7 & \textbf{85.3} & 83.1 & 83.3          \\
            grc & 74.4 & 74.8          & 71.3 & \textbf{73.7} & 69.2 & 69.6          & 68.3      & \textbf{69.0} & 64.6 & \textbf{67.3} & 62.6 & \textbf{63.4}          \\
            he  & 79.9 & 80.1          & 77.4 & \textbf{79.9} & 76.5 & \textbf{77.3} & 77.5      & 77.4          & 74.4 & \textbf{77.2} & 74.2 & 74.7          \\
            ja  & 94.2 & 94.4          & 94.2 & 94.4          & 81.3 & \textbf{81.8} & 93.2      & 93.3          & 92.7 & 93.1          & 79.5 & \textbf{80.2} \\
            zh  & 72.7 & 72.5          & 70.8 & \textbf{72.2} & 56.5 & \textbf{58.2} & 69.1      & 69.3          & 66.7 & \textbf{68.1} & 53.4 & \textbf{55.0} \\
            \midrule
            av  & 81.1 & 81.3          & 79.2 & \textbf{80.8} & 73.8 & \textbf{74.5} & 75.9      & 76.0          & 72.8 & \textbf{74.8} & 67.2 & \textbf{68.2}
        \end{tabular}
        \caption{LAS for $bi$, $bw$ and $fw$, without and with composition ($+lc$) with an LSTM. Difference $>0.5$ with $+lc$ in bold.}
        \label{tab:res}
    \end{table*}

    \indent We look at the impact of recursive compositions on all combinations of ablated models, see Table~\ref{tab:res}. We only look at the impact of using an LSTM cell rather than a recurrent cell since it was a better technique across the board (see previous section).\\
    \indent Looking first at BiLSTMs, it seems that composition does not reliably help parsing accuracy, regardless of access to POS and character information. This indicates that the vectors obtained from the \mbox{BiLSTM} already contain information that would otherwise be obtained by using composition.\\
    \indent Turning to results with either the forward or the backward LSTM ablated, we see the expected pattern. Composition helps more when the model lacks POS tags, indicating that there is some redundancy between these two methods of building contextual information. Composition helps recover a substantial part of the gap of the model with a backward LSTM with or without POS tag. It recovers a much less substantial part of the gap in other cases which means that, although there is some redundancy between these different methods of building contextual information, they are still complementary and a recursive composition function cannot fully compensate for the lack of a backward LSTM or POS and/or character information. 
    There are some language idiosyncracies in the results. While composition helps recover most of the gap for the backward LSTM models without POS and/or character information for Czech and English, it does it to a much smaller extent for Basque and Finnish. We hypothesise that arc depth might impact the usefulness of composition, since more depth means more matrix multiplications with the composition function. However, we find no correlation between average arc depth of the treebanks and usefulness of composition. It is an open question why composition helps some languages more than others.\\
    \indent Note that we are not the first to use composition over vectors obtained from a \mbox{BiLSTM} in the context of dependency parsing, as this was done by \citet{qi17arc}.
    The difference is that they compose vectors before scoring transitions. It was also done by \citet{kiperwasser16b} who showed that using \mbox{BiLSTM} vectors for words in their Tree LSTM parser is helpful but they did not compare this to using \mbox{BiLSTM} vectors without the Tree LSTM.\\
    \indent Recurrent and recursive LSTMs in the way they have been considered in this paper are two ways of constructing contextual information and making it available for local decisions in a greedy parser. The strength of recursive LSTMs is that they can build this contextual information using hierarchical context rather than linear context. A possible weakness is that this makes the model sensitive to error propagation: a wrong attachment leads to using the wrong contextual information. It is therefore possible that the benefits and drawbacks of using this method cancel each other out in the context of BiLSTMs.

    \subsection{Ensemble}
    \indent To investigate further the information captured by BiLSTMs, we ensemble the 6 versions of the models with POS and character information with the different feature extractors ($bi$, $bw$, $fw$) with ($+lc$) and without composition. We use the (unweighted) reparsing technique of \citet{sagae06naacl}\footnote{This method scores all arcs by the number of parsers predicting them and extracts a maximum spanning tree using the Chu-Liu-Edmonds algorithm \citep{edmonds67}.} 
    and ignoring labels. 
    As can be seen from the UAS scores in Table~\ref{tab:ensemble}, the ensemble (\emph{full}) largely outperforms the parser using only a BiLSTM, indicating that the information obtained from the different models is complementary. To investigate the contribution of each of the 6 models, we ablate each one by one. As can be seen from Table~\ref{tab:ensemble}, ablating either of the BiLSTM models or the backward LSTM using composition, results in the least effective of the ablated models, further strengthening the conclusion that BiLSTMs are powerful feature extractors.

    \setlength{\tabcolsep}{3pt}
    \begin{table}
        \centering
        \small
        \scalebox{0.8}{
            \begin{tabular}{l|l|l|llllll}
                & bi   & full & -bi  & -[bi+lc] & -bw  & -[bw+lc] & -fw  & -[fw+lc] \\
                \toprule
                cs   & 90.9 & 92.0  & 91.8 & 91.8   & 91.8 & 91.8   & 92.1 & 92.0   \\
                en   & 85.8 & 87.1  & 86.7 & 86.7   & 86.8 & 86.7   & 87.2 & 87.2   \\
                eu   & 78.7 & 80.9  & 80.3 & 80.2   & 80.4 & 80.3   & 80.9 & 81.0   \\
                fi   & 83.5 & 85.5  & 85.4 & 85.4   & 85.3 & 85.2   & 85.6 & 85.5   \\
                fr   & 89.8 & 90.8  & 90.8 & 90.6   & 90.8 & 90.7   & 90.8 & 90.8   \\
                grc  & 81.2 & 83.5  & 83.0 & 83.1   & 83.3 & 83.0   & 83.4 & 83.6   \\
                he   & 86.2 & 87.6  & 87.6 & 87.4   & 87.5 & 87.2   & 87.6 & 87.7   \\
                ja   & 95.9 & 96.1  & 95.8 & 95.7   & 95.9 & 95.8   & 96.3 & 96.2   \\
                zh   & 78.3 & 79.3  & 78.4 & 78.6   & 78.4 & 78.7   & 79.8 & 79.9   \\
                \midrule
                av   & 85.6 & 87.0  & 86.6 & 86.6   & 86.7 & 86.6   & 87.1 & 87.1  
            \end{tabular}
        }
        \caption{UAS ensemble (full) and ablated experiments.}
        \label{tab:ensemble}
    \end{table}

\section{Conclusion}
We investigated the impact of composing the representation of subtrees in a transition-based parser. We observed that composition does not reliably help a parser that uses a \mbox{BiLSTM} for feature extraction, indicating that vectors obtained from the \mbox{BiLSTM} might capture subtree information, which is consistent with the results of \citet{linzen16assessing}. However, we observe that, when ablating the backward LSTM, performance drops and recursive composition does not help to recover much of this gap. We hypothesise that this is because the backward LSTM primarily improves the lookahead for the greedy parser. When ablating the forward LSTM, performance drops to a smaller extent and recursive composition recovers a substantial part of the gap. This indicates that a forward LSTM and a recursive composition function capture similar information, which we take to be related to the rich history-based features crucial for a transition-based parser. To capture this information, a recursive composition function is better than a forward LSTM on its own, but it is even better to have a forward LSTM as part of a \mbox{BiLSTM}. We further find that recursive composition helps more when POS tags are ablated from the model, indicating that POS tags and a recursive composition function are partly redundant ways of constructing contextual information. Finally, we correlate results with language properties, showing that the improved lookahead of a backward LSTM is especially important for head-final languages.

\section*{Acknowledgments}

We acknowledge the computational resources provided by CSC in Helsinki and Sigma2 in Oslo through NeIC-NLPL (www.nlpl.eu). We thank Sara Stymne and Aaron Smith for many discussions about this paper.

\bibliography{expanded,main}
\bibliographystyle{acl_natbib}

\end{document}